\documentclass[10pt,twocolumn,letterpaper]{article}

\usepackage[pagenumbers]{cvpr} 

\usepackage{times}
\usepackage{epsfig}
\usepackage{graphicx}
\usepackage{amsmath}
\usepackage{amssymb}
\usepackage{mathbbol}
\usepackage{algorithmic}
\usepackage[ruled,vlined]{algorithm2e}
\usepackage{multirow}
\usepackage{multicol} 
\usepackage{array}
\usepackage{makecell}
\usepackage[dvipsnames]{xcolor}
\usepackage{booktabs}
\usepackage{arydshln}
\usepackage{subcaption} 
\usepackage{diagbox}

\usepackage[pagebackref,breaklinks,colorlinks]{hyperref}

\usepackage[capitalize]{cleveref}
\crefname{section}{Sec.}{Secs.}
\Crefname{section}{Section}{Sections}
\Crefname{table}{Table}{Tables}
\crefname{table}{Tab.}{Tabs.}

\def\cvprPaperID{2869} 
\def\confName{CVPR}
\def\confYear{2022}

\begin{document}

	\newcommand{\fracpartial}[2]{\frac{\partial #1}{\partial  #2}}
	\newcommand{\norm}[1]{\left\lVert#1\right\rVert}
	\newcommand{\innerproduct}[2]{\left\langle#1, #2\right\rangle}
	\newcommand{\fan}[1]{\Vert #1 \Vert}
	\newcommand{\qileft}{[\kern-0.15em[}
	\newcommand{\qiLeft}{\left[\kern-0.4em\left[}
	\newcommand{\qiright}{]\kern-0.15em]}
	\newcommand{\qiRight}{\right]\kern-0.4em\right]}
	\newcommand{\sign}{{\mbox{sign}}}
	\newcommand{\diag}{{\mbox{diag}}}
	\newcommand{\armin}{{\mbox{argmin}}}
	\newcommand{\rank}{{\mbox{rank}}}
	\renewcommand{\vec}{{\mbox{vec}}}
	\newcommand{\st}{{\mbox{s.t.}}}
	\newcommand{\<}{\left\langle}
	\renewcommand{\>}{\right\rangle}
	\newcommand{\lbar}{\left\|}
	\newcommand{\rbar}{\right\|}
	\renewcommand{\Roman}[1]{\uppercase\expandafter{\romannumeral#1}}
	\newcommand{\red}[1]{{\color{red}{#1}}}
	\newcommand{\blue}[1]{{\color{blue}{#1}}}

	\renewcommand{\a}{{\bm{a}}}
	\renewcommand{\b}{{\bm{b}}}
	\renewcommand{\d}{{\bm{d}}}
	\newcommand{\e}{{\bm{e}}}
	\newcommand{\f}{{\bm{f}}}
	\newcommand{\g}{{\bm{g}}}
	\renewcommand{\o}{{\bm{o}}}
	\newcommand{\p}{{\bm{p}}}
	\newcommand{\q}{{\bm{q}}}
	\renewcommand{\r}{{\bm{r}}}
	\newcommand{\s}{{\bm{s}}}
	\renewcommand{\t}{{\bm{t}}}
	\renewcommand{\u}{{\bm{u}}}
	\renewcommand{\v}{{\bm{v}}}
	\newcommand{\w}{{\bm{w}}}
	\newcommand{\x}{{\bm{x}}}
	\newcommand{\y}{{\bm{y}}}
	\newcommand{\z}{{\bm{z}}}
	\newcommand{\balpha}{{\bm{\alpha}}}
	\newcommand{\bbeta}{{\bm{\beta}}}
	\newcommand{\bmu}{{\bm{\mu}}}
	\newcommand{\bsigma}{{\bm{\sigma}}}
	\newcommand{\blambda}{{\bm{\lambda}}}
	\newcommand{\btheta}{{\bm{\theta}}}
	\newcommand{\bgamma}{{\bm{\gamma}}}
	\newcommand{\bxi}{{\bm{\xi}}}
	\newcommand{\bphi}{{\bm{\phi}}}
	
	\newcommand{\ba}{{\bm{A}}}
	\newcommand{\bb}{{\bm{B}}}
	\newcommand{\bc}{{\bm{C}}}
	\newcommand{\bd}{{\bm{D}}}
	\newcommand{\be}{{\bm{E}}}
	\newcommand{\bg}{{\bm{G}}}
	\newcommand{\bi}{{\bm{I}}}
	\newcommand{\bj}{{\bm{J}}}
	\newcommand{\bl}{{\bm{L}}}
	\newcommand{\bo}{{\bm{O}}}
	\newcommand{\bp}{{\bm{P}}}
	\newcommand{\bq}{{\bm{Q}}}
	\newcommand{\bs}{{\bm{S}}}
	\newcommand{\bu}{{\bm{U}}}
	\newcommand{\bv}{{\bm{V}}}
	\newcommand{\bw}{{\bm{W}}}
	\newcommand{\bx}{{\bm{X}}}
	\newcommand{\by}{{\bm{Y}}}
	\newcommand{\bz}{{\bm{Z}}}
	\newcommand{\bTheta}{{\bm{\Theta}}}
	\newcommand{\bSigma}{{\bm{\Sigma}}}
	
	\newcommand{\A}{{\mathcal{A}}}
	\newcommand{\B}{\mathcal{B}}
	\newcommand{\C}{\mathcal{C}}
	\newcommand{\D}{\mathcal{D}}
	\newcommand{\F}{\mathcal{F}}
	\renewcommand{\H}{\mathcal{H}}
	\newcommand{\I}{\mathcal{I}}
	\renewcommand{\L}{\mathcal{L}}
	\newcommand{\N}{\mathcal{N}}
	\renewcommand{\P}{\mathcal{P}}
	\newcommand{\X}{\mathcal{X}}
	\newcommand{\Y}{\mathcal{Y}}
	\newcommand{\W}{\mathcal{W}}

\title{SimMatch: Semi-supervised Learning with Similarity Matching \vspace{-5pt}}


\author{%
	Mingkai Zheng${}^{1,2}$ \quad Shan You$^{2}$\thanks{Correspondence to: Shan You $<$\texttt{youshan@sensetime.com}$>$.} \\ \quad Lang Huang$^3$ \quad Fei Wang$^4$ \quad Chen Qian$^2$ \quad Chang Xu$^1$\\
	\normalsize $^1$School of Computer Science, Faculty of Engineering, The University of Sydney \\ 
	\normalsize $^2$SenseTime Research\quad
	\normalsize $^3$ The University of Tokyo  \\
	\normalsize $^4$University of Science and Technology of China \\
}

\maketitle

\begin{abstract}
Learning with few labeled data has been a longstanding problem in the computer vision and machine learning research community. In this paper, we introduced a new semi-supervised learning framework, SimMatch, which simultaneously considers semantic similarity and instance similarity. In SimMatch, the consistency regularization will be applied on both semantic-level and instance-level. The different augmented views of the same instance are encouraged to have the same class prediction and similar similarity relationship respected to other instances. Next, we instantiated a labeled memory buffer to fully leverage the ground truth labels on instance-level and bridge the gaps between the semantic and instance similarities. Finally, we proposed the \textit{unfolding}  and \textit{aggregation} operation which allows these two similarities be isomorphically transformed with each other. In this way, the semantic and instance pseudo-labels can be mutually propagated to generate more high-quality and reliable matching targets.   Extensive experimental results demonstrate that SimMatch improves the performance of semi-supervised learning tasks across different benchmark datasets and different settings. Notably, with 400 epochs of training, SimMatch achieves 67.2\%, and 74.4\% Top-1 Accuracy with 1\% and 10\% labeled examples on ImageNet, which significantly outperforms the baseline methods and is better than previous semi-supervised learning frameworks. Code and pre-trained models are available at \href{https://github.com/KyleZheng1997/simmatch}{https://github.com/KyleZheng1997/simmatch}
\vspace{-15pt}
\end{abstract}

\section{Introduction}
Benefiting from the availability of large-scale annotated datasets and growing computational resources in the last decades, deep neural networks have demonstrated their success on a variety of visual tasks \cite{imagenet_cvpr09, coco, pascal-voc-2007,  prioritized, greedynas, maskrcnn, fastrcnn}. However, a large volume of labeled data is very expensive to collect in a real-world scenario. Learning with few labeled data has been a longstanding problem in the computer vision and machine learning research community. Among various methods, semi-supervised learning (SSL) \cite{introductionSSL, continuationSSL, surveySSL, survey2SSL} serves as an effective solution by dint of the help of massive unlabeled data, and achieves remarkable performance. 

\begin{figure}
    \centering
    \includegraphics[width=0.8\linewidth]{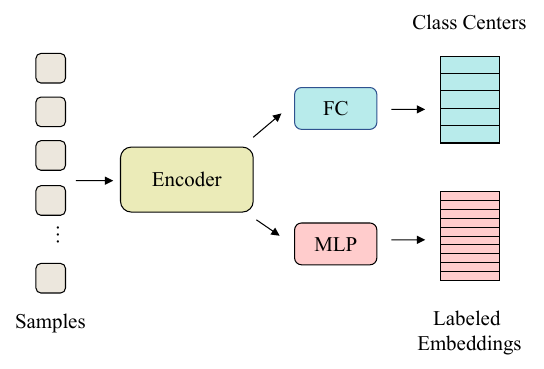}
    \vspace{-10pt}
    \caption{A sketch of  SimMatch. The Fully-Connected layer vectors can be viewed as a semantic representative or class center for each category. However, due to the limited labeled samples, the semantic-level information is not always reliable. In SimMatch, we consider the instance-level and semantic-level information simultaneously and adopt a labeled memory buffer to fully leverage the ground truth label on instance-level.}
    \vspace{-15pt}
    \label{fig:motivation}
\end{figure}

A simple but very effective semi-supervised learning method is to pretrain a model on a large-scale dataset and then transfer the learned representation by fine-tuning the pretrained model with a few labeled samples. Thanks to the recent advances in self-supervised learning \cite{moco, simclr, simclrv2, byol, swav, mediantriplet, nnclr}, such pretraining and fine-tuning pipeline have demonstrated its promising performance in SSL. Most self-supervised learning frameworks focus on the design of pretext tasks. For example, instance discrimination \cite{instance_discrimination} encourages different views of the same instance to share the same features, and different instances should have distinct features. Deep clustering based methods \cite{deepclustering, swav, self_label} expect different augmented views of the same instance should be classified into the same cluster. However, most of these pretext tasks are designed in a completely unsupervised manner, without considering the few labeled data at hand.

Instead of standalone two-stage pretraining and fine-tuning, current popular methods directly involve the labeled data in a joint feature learning paradigm with pseudo-labeling \cite{pseudolabel} or consistency regularization \cite{regularization}. The main idea behind these methods is to train a semantic classifier with labeled samples and use the predicted distribution as the pseudo label for the unlabeled samples. In this way, the pseudo-labels are generally produced by the weakly augmented views \cite{remixmatch, fixmatch} or the averaged predictions of multiple strongly augmented views \cite{mixmatch}. The objective will be constructed by the cross entropy loss between an different strongly augmented views and the pseudo-labels. It might also be noted that the pseudo-labels will generally be sharpened or operated by $argmax$ since every instance is expected to be classified into a category. However, when there are only very limited annotated data, the semantic classifier is no longer reliable; applying the pseudo-label method will cause the ``overconfidence" issue \cite{co-semi-training, zhang2021understanding}, which means the model will fit on the confident but wrong pseudo-labels, resulting in poor performance.

In this paper, we introduce a novel semi-supervised learning framework, SimMatch, which has been shown in Figure \ref{fig:motivation}. In SimMatch, we bridge both sides and propose to match the similarity relationships of both semantic and instance levels simultaneously for different augmentations. Specifically, we first require the strongly augmented view to have the same semantic similarity (\ie label prediction) with a weak augmented view; besides, we also encourage the strong augmentation to have the same instance characteristics (\ie similarity between instances) with the weak one for more intrinsic feature matching.  Moreover, different from the previous works that simply regard the predictions of the weakly augmented views as pseudo-labels. In SimMatch, the semantic and instance pseudo-labels are allowed to interact by instantiating a memory buffer that keeps all the labeled examples. In this way, these two similarities can be isomorphically transformed with each other by introducing \textit{aggregating} and \textit{unfolding} techniques. Thus the semantic and instance pseudo-labels can be mutually propagated to generate more high-quality and reliable matching targets. Extensive experiments demonstrate the effectiveness of SimMatch across different settings. Our contribution can be summarized as follows:

\begin{itemize}
 \vspace*{-0.3em}
 \setlength\itemsep{-0.1em}
 \item We proposed SimMatch, a novel semi-supervised learning framework that simultaneously considers semantics similarity and instance similarity. 
 
 \item To channel both similarities, we leverage a labeled memory buffer so that semantic and instance pseudo-labels can be mutually propagated with the \textit{aggregating} and \textit{unfolding} techniques.

 \item SimMatch establishes a new state-of-the-art performance for semi-supervised learning. With only 400 epochs of training, SimMatch achieves 67.2\% and 74.4\% Top-1 accuracy with 1\% and 10\% labeled examples on ImageNet. 
 
\end{itemize}

\section{Related Work}

\subsection{Semi-Supervised Learning}
Consistency Regularization is a widely adopted method in semi-supervised learning. The main idea is to enforce the model to output a consistent prediction for the different perturbed versions of the same instance. For example,  \cite{temporal, regularization} achieved such consistency requirement by minimizing the mean square difference between the predicted probability distribution of the two transformed views. In this case, the transformation could be either domain-specific data augmentations \cite{mixmatch, remixmatch, fixmatch},  or some regularization techniques in the network (\eg drop out \cite{dropout} and random max-pooling \cite{regularization}). Moreover, \cite{temporal} also proposed a temporal ensemble strategy to aggregate the predictions of multiple previous networks, which makes the predicted distribution more reliable. Mean Teacher \cite{meanteacher} further extends this idea which replaced the aggregated predictions with the output of an exponential moving average (EMA) model. 

MixMatch \cite{mixmatch}, ReMixMatch \cite{remixmatch}, and FixMatch \cite{fixmatch} are three augmentation anchoring based methods that fully leverage the augmentation consistency. Specifically, MixMatch adopts a sharpened averaged prediction of multiple strongly augmented views as the pseudo label and utilizes the MixUp trick \cite{mixup} to further enhance the pseudo label. ReMixMatch improved this idea by generating the pseudo label with weakly augmented views and also introduced a distribution alignment strategy that encourages the pseudo label distribution to match the marginal distribution of ground-truth class labels. FixMatch simplified these ideas, where the unlabeled images are only retained if the model produces a high-confidence pseudo label. Despite its simplicity, FixMatch achieved state-of-the-art performance among the augmentation anchoring-based methods.

\begin{figure*}[!h]
    \centering
    \includegraphics[width=0.8\linewidth]{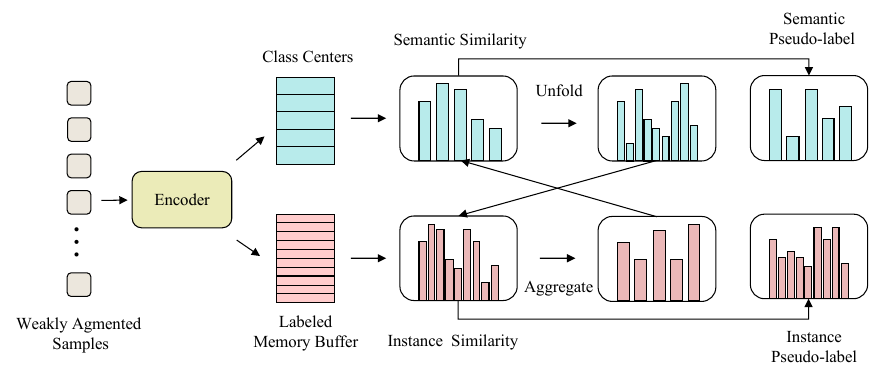}
    \vspace{-15pt}
    \caption{An overview of the SimMatch pseudo-label generation process. SimMatch will use the weak augmented view to generate a semantic pseudo-label and an instance pseudo-label. Specifically, we will first compute the semantic and instance similarity by the class centers and labeled embeddings, then use the unfolding and aggregation operations to fuse these two similarities and finally get the pseudo-label. Please see more details in our method section below. }
    \vspace{-15pt}
    \label{fig:framework}
\end{figure*}

\subsection{Self-supervised Pretraining}
Apart from the typical semi-supervised learning method, self-supervised and contrastive learning \cite{instance_discrimination, simclr, moco} has gained much attention in this research community since fine-tuning the pre-trained model with labeled samples has shown promising classification results, especially SimCLR v2 \cite{simclrv2} shows that a big (deep and wide) pre-trained model is a strong semi-supervised learner.  Most of the contrastive learning framework adopts the instance discrimination \cite{instance_discrimination} as the pretext task, which defines different augmented views of the same instance as positive pairs, while negative pairs are formed by sampling views from different instances. However, because of the existence of similar samples, treating different instances as negative pairs will result in a class collision problem \cite{contrastive_theory}, which is not conducive to the downstream tasks (especially classification tasks). Some previous works addressed this issue by unsupervised clustering \cite{deepclustering, PCL, WCL, swav}, where similar samples will be clustered into the same class. There are also some other methods designed various negative free pretext tasks \cite{byol, SimSiam, adco, ressl, msf} to avoid the class collision problem. Both cluster-based methods and negative-free based methods have shown significant improvements for downstream classification tasks.

CoMatch \cite{comatch} combines the idea of consistency regularization and contrastive learning, where the target similarity of two instances is measured by the similarity between two class probability distributions, which achieves the current state-of-the-art performance on semi-supervised learning. However, it is very sensitive to the hyper-parameters, the optimal temperature and threshold is different for various datasets and settings. Compared to CoMatch, SimMatch is faster, more robust, and has higher performance. 

\section{Method}
In this section, we will first revisit the preliminary work on augmentation anchoring based semi-supervised learning frameworks; then, we will introduce our proposed method SimMatch. After that, the algorithm and the implementation details will also be explained. 

\subsection{Preliminaries}
We define the semi-supervised image classification problem as following. Given a batch of $B$ labeled samples $ \mathcal{X}  = \{ x_b : b \in (1,...,B) \}$, we randomly apply a weak augmentation function (\eg using only a flip and a crop) $T_w(\cdot)$ to obtain the weakly augmented samples. Then, a convolutional neural network based encoder $F(\cdot)$ is employed to extract the feature information from these samples, \ie $\mathbf{h} = \mathcal{F}(T(x))$. Finally, a fully connected class prediction head $\phi(\cdot)$ is utilized to map $\mathbf{h}_b$ into semantic similarities, which can be written as: $p = \phi( \mathbf{h} )$. The labeled samples could be directly optimized by the cross entropy loss with the ground truth labels:
\begin{equation}
    \label{equation:loss_s}
    \mathcal{L}_{s} = \frac{1}{B} \sum \text{H}(y, p)
\end{equation}

Let us define a batch of $\mu B$ unlabeled samples $\mathcal{U} = \{ u_b : b \in (1,...,\mu B) \} $. By following \cite{remixmatch, mixmatch}, we randomly apply the weak and strong augmentation $T_w(\cdot)$, $T_s(\cdot)$ and using the same processing step as the labeled samples to get the semantic similarities for weakly augmented sample $p^w$ (pseudo label) and strongly augmented sample $p^s$. Then the unsupervised classification loss can be defined as the cross-entropy between these two predictions:
\begin{equation}
    \label{equation:loss_u}
    \mathcal{L}_{u} = \frac{1}{\mu B} \sum \mathbb{1} (\max DA(p^w) > \tau ) \text{H}(DA(p^w), p^s)
\end{equation}

Where $\tau$ is the confidence threshold. Following \cite{fixmatch}, we only retain the unlabeled samples whose largest class probability in the pseudo-labels are larger than $\tau$. $DA(\cdot)$ stands for the distribution alignment strategy from \cite{remixmatch} which balanced the pseudo-labels distribution. We simply follow the implementation from \cite{comatch} where we maintain a moving-average of $p^w_{avg}$ and adjust the current $p^w$ with  $Normalize(p^w / p^w_{avg})$.  Please also noted that we do not take the sharpened or one-hot version of $p^w$, $DA(p^w)$ will be directly served as the pseudo-label.

\subsection{Instance Similarity Matching}
In SimMatch, we also consider the instance-level similarity as we have discussed previously. Concretely, we encourage the strongly augmented view to have a similar similarity distribution with the weakly augmented view. Suppose we have a non-linear projection head $g(\cdot)$ which maps the representation $\mathbf{h}$ to a low-dimensional embedding $\mathbf{z}_b = g(\mathbf{h}_b)$. Following the anchoring based method, we use $\mathbf{z}^w_b$ and $\mathbf{z}^s_b$ to denote the embedding from the weakly and strongly augmented view. Now, lets assume we have $K$ weakly augmented embeddings for a bunch of different samples $\{\mathbf{z}_k: k \in (1,...,K) \}$, we calculate the similarities between $\mathbf{z}^w$ and $i$-th instance by using a similarity function $sim(\cdot)$, which represents the dot product between $L_2$ normalized vectors $sim(\mathbf{u}, \mathbf{v}) = \mathbf{u}^{T}\mathbf{v} / \lVert \mathbf{u} \lVert \lVert \mathbf{v} \lVert$. A softmax layer can be adopted to process the calculated similarities, which then produces a distribution:
\begin{align}
    q^w_i &= \frac{\exp(sim(\mathbf{z}^w_b, \mathbf{z}_{i})/ t) }{\sum_{k=1}^{K}  \exp(sim(\mathbf{z}^w_b, \mathbf{z}_{k}) / t) } \label{equation:teacher}
\end{align}

Where $t$ is the temperature parameter that controls the sharpness of the distribution. On the other hand,  we can calculate the similarities between the strongly augmented view $\mathbf{z}^s$ and $\mathbf{z}_i$ as $sim(\mathbf{z}^s_b, \mathbf{z}_i)$. The resulting similarity distribution can be written as:
\begin{align}
    q^s_i &= \frac{\exp(sim(\mathbf{z}^s_b, \mathbf{z}_{i})/ t) }{\sum_{k=1}^{K}  \exp(sim(\mathbf{z}^s_b, \mathbf{z}_{k}) / t) } \label{equation:student}
\end{align}

Finally, the consistency regularization can be achieved by minimizing the different between $q^s$ and $q^w$. Here, we adopt the cross entropy loss, which can be formulated as:
\begin{equation}
    \label{equation:loss_instance}
    \mathcal{L}_{in} = \frac{1}{\mu B} \sum \text{H}(q^w, q^s)
\end{equation}

Please noted that the instance consistency regularization will only be applied on the unlabeled examples.  The overall training objective for our model will be:
\begin{equation}
    \label{equation:overall}
    \mathcal{L}_{overall} = \mathcal{L}_{s} + \lambda_u \mathcal{L}_u  + \lambda_{in} \mathcal{L}_{in}
\end{equation}
where $\lambda_u$ and $\lambda_{in}$ are the balancing factors that control the weights of the two losses.

\subsection{Label Propagation through SimMatch}
Although our overall training objective also considers the consistency regularization on instance-level, however, the generation of the instance pseudo-labels $q^w$ are still in a fully unsupervised manner, which is absolutely a waste of the labeled information. To improve the quality of the pseudo-labels, in this section, we will illustrate how to leverage the labeled information on instance-level and also introduce a way that allows semantic similarity and instance similarity to interact with each other. 

We instantiated a labeled memory buffer to keep all the annotated examples as we have shown in Figure \ref{fig:framework} (red branch). In this way, each $\mathbf{z}_k$ that we used in Eq.\eqref{equation:teacher} and Eq.\eqref{equation:student} could be assigned to a specific class. If we interpret the vectors in $\phi$ as the ``centered" class references, the embeddings in our labeled memory buffer can be viewed as a set of ``individual" class references. 

By given a weakly augmented sample, we first compute the semantic similarity $p^w \in \mathbf{R}^{1\times L} $ and instance similarity $q^w \in \mathbf{R}^{1\times K}$.  (Noted that $L$ is generally much smaller than $K$ since we need at least one sample for each class.)  To calibrate $q^w$ with $p^w$, we need to \textbf{unfold} $p^w$ into $K$ dimensional space which we denote it as $p^{unfold}$. we achieved this by matching the corresponding semantic similarity for each labeled embedding:
\begin{equation}
    \label{equation:unfold}
    p^{unfold}_{i} = p^w_j, \; where \;\; class(q^w_j) =  class(p^w_i)
\end{equation}

where $class(\cdot)$ is the function that returns the ground truth class. Specifically, $class(q^w_j)$ represent the label for the $j^{th}$ element in memory buffer and $class(p^w_i)$ means the $i^{th}$ class. Now, we regenerate the calibrated instance pseudo-labels by \textbf{scaling} $q^w$ with $p^{unfold}$, which can be expressed as the following:
\begin{equation}
    \label{equation:calibrateq}
    \widehat{q}_{i} = \frac{q^w_i p^{unfold}_i}{\sum_{k=1}^{K} q^w_k p^{unfold}_k } 
\end{equation}

The calibrated instance pseudo-label $\widehat{q}$ will be served as a new target and replace the old one $q^w$ in Eq.\eqref{equation:loss_instance}. On the other hand, we can also use the instance similarity to adjust the semantic similarity. To do this, we first need to \textbf{aggregate} $q$ into $L$ dimensional space which we denote it as $q^{agg}$. We achieved this by sum over the instance similarities that share the same ground truth labels:
\begin{equation}
    \label{equation:aggregation}
    q^{agg}_{i} = \sum_{j=0}^{K} \mathbb{1} (class(p^w_i) = class(q^w_j)) q^w_j
\end{equation}

\begin{figure}
    \centering
    \includegraphics[width=0.9\linewidth]{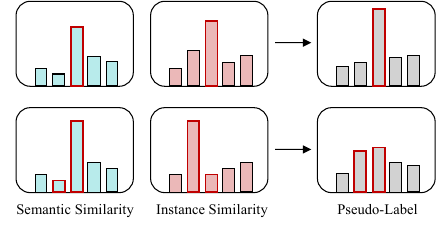}
    \vspace{-3mm}
    \caption{The intuition behind label propagation. If the semantic and instance similarities are similar, the result pseudo-label will be much sharper and produce high confidence for some classes. When these two similarities are different, the result pseudo-label will be much flatter.}
    \label{fig:fuse}
    \vspace{-5mm}
\end{figure}

Now, we regenerate the adjusted semantic pseudo-label by \textbf{smoothing} $p^w$ with $q^{agg}$, which can be written as:

\begin{equation}
    \label{equation:calibratep}
    \widehat{p}_i = \alpha p^w_i + (1 - \alpha) q^{agg}_{i}
\end{equation}
where $\alpha$ is the hyper-parameter that controls the weight of the semantic and instance information. Similarly, The adjusted semantic pseudo-label will replace the old one $p^w_i$ in Eq.\eqref{equation:loss_u}.  In this way, the pseudo-label $\widehat{p}$ and $\widehat{q}$ will both contains the semantic-level and instance-level information.  As we have shown in Figure \ref{fig:fuse}, when semantic and instance similarities are similar, which means these two distributions agree with the prediction of each other, then the result pseudo-label will be much sharper and produce high confidence for some classes. On the other hand, if these two similarities are different, the result pseudo-label will be much flatter and not contain high probability values. In SimMatch, we adopt the scaling and smoothing strategy for $\widehat{q}$ and $\widehat{p}$ respectively, we also have tried different combination for these two strategies, please see more details in our ablation study section.

\subsection{Efficient Memory Buffer}
As we have mentioned, SimMatch requires a memory buffer to keep the embeddings for labeled examples. In doing so, we are required to store both feature embeddings and the ground truth labels. Specifically, we defined a feature memory buffer $Q_f \in \mathbf{R}^{K \times D}$  and a label memory buffer $Q_l \in \mathbf{R}^{K \times 1}$ where $K$ is the number of available annotated samples, $D$ is the embedding size. The largest $K$ in our experiments is around $10^5$ (ImageNet 10\% setting), which only costs 64M GPU memories for $Q_f$. For $Q_l$, we just need to store a scalar for each label, the aggregation and unfolding operation can be easily achieved by $gather$ and $scatter\_add$ function, which should has been efficiently implemented in recent deep learning libraries \cite{pytorch, tensorflow}. In this case, $Q_l$ only costs less than 1M GPU memories ($K=10^5$), which is almost negligible.

According to \cite{moco}, the rapid changed feature in memory buffer will dramatically reduce the performance. In SimMatch, we adopt two different implementation for different buffer size. When $K$ is large, we follow MoCo \cite{moco} to leverage a student-teacher based framework, we denote it as $\mathcal{F}_s$ and $\mathcal{F}_t$. In this case, the labeled examples and strongly augmented samples will be passed into $\mathcal{F}_s$, the weakly augmented samples will be feed into $\mathcal{F}_t$ to generate the pseudo-labels. The parameters of $\mathcal{F}_t$ will be updated by:
\begin{equation} \label{equation:ema}
    \mathcal{F}_t \leftarrow m \mathcal{F}_t + (1 - m)\mathcal{F}_s
\end{equation}

On the other hand, when $K$ is small, maintain a teacher network is not necessary, We simply adopt the temporal ensemble strategy \cite{temporal, instance_discrimination} to smooth the features in memory buffer, which can be written as :
\begin{equation} \label{equation:ema}
    \mathbf{z}_t \leftarrow m \mathbf{z}_{t-1} + (1 - m)\mathbf{z}_{t}
\end{equation}
In this case, all the samples will be directly pass into the same encoder. The student-teacher version of SimMatch has been illustrated in Algorithm \ref{alg:simmatch}.

\begin{algorithm}[t]
\SetAlgoLined
\SetKwInOut{Input}{Input}
\Input{$\mathbf{x}_l$ and :$\mathbf{x}_u$  a batch of labeled and unlabeled samples. $T_w(\cdot)$ and $T_s(\cdot)$: Weak and strong augmentation function. $\mathcal{F}_t$ and $\mathcal{F}_s$: Teacher and student encoder. $\phi_t$ and $\phi_s$: teacher and student classifier. $g_t$ and $g_s$: teacher and student projection head. $Q_f$ and $Q_l$: The feature and label memory buffer.} 

\While{network not converge} {
    \For{i=1 to step}{
        $\mathbf{h}^w$ = $\mathcal{F}_t(T_w(\mathbf{x}_u)) \quad\quad \mathbf{h}^s = \mathcal{F}_s(T_s(\mathbf{x}_u))$ \\
        $p^w = DA(\phi_t(\mathbf{h}^w))  \quad\;\;   p^s = \phi_s(\mathbf{h}^s)$      \\
        $\mathbf{z}^w$ = $g_t(\mathbf{h}^w) \quad\quad\quad\quad\;  \mathbf{z}^s = g_s(\mathbf{h}^s)$ \\
        $\mathbf{h}^l_t$ = $\mathcal{F}_t(T_w(\mathbf{x}_l)) \quad\quad\;\; \mathbf{h}^l_s = \mathcal{F}_s(T_w(\mathbf{x}_l))$ \\
        $p^l = \phi_s(\mathbf{h}^l_s) \quad\quad\quad\quad\;\; \mathbf{z}^l = g_t(\mathbf{h}^l_t)$ \\
        Compute $q^w$ and $q^s$ by Eq.\eqref{equation:teacher} Eq.\eqref{equation:student} \\
        Compute $p^{unfold}$ and $q^{agg}$ by Eq.\eqref{equation:unfold} Eq.\eqref{equation:aggregation} \\
        Compute $\widehat{q}$ and $\widehat{p}$ by Eq.\eqref{equation:calibrateq} Eq.\eqref{equation:calibratep} \\
        $\mathcal{L}_{s} = \frac{1}{B} \sum \text{H}(y, p^l)$ \\
        $\mathcal{L}_{u} = \frac{1}{\mu B} \sum \mathbb{1} (\max \widehat{p} > \tau ) \text{H}(\widehat{p}, p^s)$ \\
        $\mathcal{L}_{in} = \frac{1}{\mu B} \sum \text{H}(\widehat{q}, q^s)$ \\
        $\mathcal{L}_{overall} = \mathcal{L}_s + \lambda_u \mathcal{L}_u  + \lambda_{in} \mathcal{L}_{in} $ \\
        Optimize $F_s$, $g_s$ and $\phi_s$ by $\mathcal{L}_{overall}$ \\
        Momentum Update $F_t$, $g_t$ and $\phi_t$ \\
        Update $Q_f$ and $Q_l$ with $\mathbf{z}^l$ and $y$ 
    }
}
\SetKwInOut{Output}{Output}
\Output{The well trained model $\mathcal{F}_s$ and $g_s$}
\caption{SimMatch (Student-Teacher)}
\label{alg:simmatch}
\end{algorithm}

\section{Experiments}
In this section, we will first test SimMatch on various dataset and settings to shows its superiority, then we will ablation each component to validate the effectiveness of each component in our framework.

\subsection{CIFAR-10 and CIFAR-100}
We first evaluation SimMatch on CIFAR-10 and CIFAR-100 \cite{cifar} datasets. CIFAR-10 consists of 60000 32x32 color images in 10 classes, with 6000 images per class. There are 50000 training images and 10000 test images. CIFAR-100 is just like the CIFAR-10, except it has 100 classes containing 600 images each. There are 500 training images and 100 testing images per class. For CIFAR-10, we randomly sample 4, 25, and 400 samples from the training set as the labeled data and use the rest of the training set as the unlabeled data. For CIFAR-100, we perform the same experiments but use 4, 25, and 100 samples per class.

\renewcommand\arraystretch{0.8}
\begin{table*}
    \centering
    \setlength\tabcolsep{8pt}
    \caption{Top-1 Accuracy comparison (mean and std over 5 runs) on CIFAR-10 and CIFAR-100 with varying labeled set sizes.}
    \vspace{-5pt}
    \small
    \begin{tabular}{lrrrrrr}
    \toprule
        & \multicolumn{3}{c}{CIFAR-10} & \multicolumn{3}{c}{CIFAR-100} \\
        \cmidrule(l{3pt}r{3pt}){1-1} \cmidrule(l{3pt}r{3pt}){2-4}  \cmidrule(l{3pt}r{3pt}){5-7}
        Method & 40 labels & 250 labels & 4000 labels & 400 labels & 2500 labels & 10000 labels \\
        \cmidrule(l{3pt}r{3pt}){1-1} \cmidrule(l{3pt}r{3pt}){2-4}  \cmidrule(l{3pt}r{3pt}){5-7}
        $\Pi$-Model \cite{temporal} &  - &  45.74$\pm$3.97 & 58.99$\pm$0.38 &  - & 42.75$\pm$0.48 & 62.12$\pm$0.11 \\
        Pseudo-Labeling \cite{pseudolabel} &  - &  50.22$\pm$0.43 & 83.91$\pm$0.28 &  - & 42.62$\pm$0.46 & 63.79$\pm$0.19 \\
        Mean Teacher \cite{meanteacher} &  - &  67.68$\pm$2.30 & 90.81$\pm$0.19 &  - & 46.09$\pm$0.57 & 64.17$\pm$0.24 \\
        UDA \cite{uda} &  70.95$\pm$5.93 &  91.18$\pm$1.08 & 95.12$\pm$0.18 & 40.72$\pm$0.88  & 66.87$\pm$0.22 & 75.50$\pm$0.25 \\
        MixMatch \cite{mixmatch} &  52.46$\pm$11.50 &  88.95$\pm$0.86 & 93.58$\pm$0.10 & 32.39$\pm$1.32  & 60.06$\pm$0.37 & 71.69$\pm$0.33 \\
        ReMixMatch \cite{remixmatch} &  80.90$\pm$9.64 &  94.56$\pm$0.05 & 95.28$\pm$0.13 & 55.72$\pm$2.06  & 72.57$\pm$0.31 & 76.97$\pm$0.56 \\
        FixMatch(RA) \cite{fixmatch} &  86.19$\pm$3.37 &  94.93$\pm$0.65 & 95.74$\pm$0.05 & 51.15$\pm$1.75  & 71.71$\pm$0.11 & 77.40$\pm$0.12 \\
        Dash \cite{dash} &  86.78$\pm$3.75 &  \textbf{95.44}$\pm$0.13 & 95.92$\pm$0.06 & 55.24$\pm$0.96  & 72.82$\pm$0.21 & 78.03$\pm$0.14 \\
        CoMatch \cite{comatch} &  93.09$\pm$1.39 &  95.09$\pm$0.33 & - & -  & - & - \\
        \cmidrule(l{3pt}r{3pt}){1-1} \cmidrule(l{3pt}r{3pt}){2-4}  \cmidrule(l{3pt}r{3pt}){5-7}
        
        SimMatch(Ours) &  \textbf{94.40}$\pm$1.37 & 95.16$\pm$0.39  & \textbf{96.04}$\pm$0.01 & \textbf{62.19}$\pm$2.21  & \textbf{74.93}$\pm$0.32 & \textbf{79.42}$\pm$0.11  \\
        \bottomrule
    \end{tabular}
    \vspace{-8pt}
    \label{table:cifar_result}
\end{table*}

\textbf{Implementation Details}. Most our implementations follows \cite{fixmatch}. Specifically, we adopt WRN28-2, and WRN28-8 \cite{wrn} for CIFAR-10 and CIFAR-100 respectively. We use a standard SGD optimizer with Nesterov momentum \cite{sgdmomentum, polyak1964some} and set the initial learning rate to 0.03. For the learning rate schedule, we use a cosine learning rate decay \cite{cosine_lr} which adjust the learning rate to $0.03 \cdot cos(\frac{7\pi s}{16S})$ where $s$ is the current training step, and $S$ is the total number of training steps. We also report the final performance using an exponential moving average of model parameters. Note that we use an identical set of hyper-parameters for both datasets ($\lambda_u=1, \lambda_{in}=1, t=0.1, \alpha=0.9, \tau=0.95, \mu=7, m=0.7, B=64, S=2^{20}$) . For distribution alignment, we accumulate the past 32 steps $p^w$ for calculating the moving average $p^w_{avg}$. We adopt the temporal ensemble memory buffer \cite{temporal} since most settings for these two datasets have a relatively small $K$. For the implementations of the strong and weak augmentations, we strictly follow the FixMatch \cite{fixmatch}.

\textbf{Results}. The result has been reported in table \ref{table:cifar_result}. For baseline, we mainly consider methods $\Pi$-Model \cite{temporal}, Pseudo-Labeling \cite{pseudolabel}, Mean Teacher \cite{meanteacher}, UDA \cite{uda}, MixMatch \cite{mixmatch}, ReMixMatch \cite{remixmatch}, FixMatch \cite{fixmatch}, and CoMatch \cite{comatch}. We compute the mean and variance of accuracy when training on 5 different “folds” of labeled data. As we can see that the SimMatch achieves state-of-the-art performance on various settings, especially on CIFAR-100. For CIFAR-10, SimMatch has a large performance gain on 40 labels setting, but the improvements for 250 and 4000 is relatively small. We doubt that this is due to the accuracy of $95\% \sim 96\%$ being already quite close to the supervised performance.

\renewcommand\arraystretch{0.8}
\begin{table*}
    \small
	\centering
	\caption{Experimental results on ImageNet with 1\% and 10\% labeled examples.}
	\vspace{-8pt}
	\begin{tabular}	{l | l | l | l | c  c |c  c }
	\toprule
	\multirow{3}{*}{\shortstack[l]{Self-supervised\\Pre-training}} & \multirow{3}{*}{Method} & \multirow{3}{*}{Epochs} & \multirow{3}{*}{\shortstack[l]{Paramters\\ (train/test)}} & \multicolumn{2}{c|}{Top-1} & \multicolumn{2}{c}{Top-5} \\
	& & &    & \multicolumn{2}{c|}{Label fraction} & \multicolumn{2}{c}{Label fraction} \\
	 & & &    &  1\% & 10\%& 1\% & 10\%\\
	\midrule
	\multirow{7}{*}{None} &Pseudo-label~\cite{pseudolabel,S4L} & $\sim$100 & 25.6M / 25.6M & - & - & 51.6&82.4\\ 
	&VAT+EntMin.~\cite{VAT,EntMin,S4L}& - & 25.6M / 25.6M & - & 68.8 & -&88.5\\ 
		& S4L-Rotation~\cite{S4L} & $\sim$200 & 25.6M / 25.6M & - & 53.4 & - &83.8\\ 
		& UDA ~\cite{uda} & - &  25.6M / 25.6M & - & 68.8 & - &88.5\\
	 & FixMatch ~\cite{fixmatch} & $\sim$300 &  25.6M / 25.6M & - & 71.5 & - & 89.1\\
	& CoMatch \cite{comatch} &  $\sim$400  &  30.0M / 25.6M & 66.0 & 73.6 & 86.4 & \textbf{91.6} \\ 
	\midrule
	PCL~\cite{PCL} & \multirow{5}{*}{Fine-tune} &$\sim$200  &  25.8M / 25.6M  & - & - & 75.3&85.6 \\ 
	SimCLR~\cite{simclr} & &$\sim$1000  &  30.0M / 25.6M  & 48.3& 65.6  & 75.5&87.8 \\ 
	SimCLR V2~\cite{simclrv2} & &$\sim$800  &  34.2M / 25.6M  & 57.9& 68.4  & 82.5 &89.2 \\ 
	BYOL~\cite{byol} & &$\sim$1000 &  37.1M / 25.6M  & 53.2&  68.8& 78.4&   89.0\\ 
	SwAV~\cite{swav} & &$\sim$800 &  30.4M / 25.6M  & 53.9 & 70.2 &  78.5 &   89.9\\ 
	WCL~\cite{WCL} & &$\sim$800 &  34.2M / 25.6M  & 65.0 & 72.0 &  86.3 & 91.2\\
	\cmidrule{2-8}
	\multirow{3}{*}{MoCo V2 ~\cite{mocov2}}  & Fine-tune&$\sim$800  &  30.0M / 25.6M  & 49.8 & 66.1& 77.2 & 87.9\\ 
	&CoMatch \cite{comatch} & $\sim$1200 & 30.0M / 25.6M & 67.1 & 73.7 & \textbf{87.1} & 91.4\\
	\midrule
	MoCo-EMAN \cite{eman} &FixMatch-EMAN \cite{eman} & $\sim$1100 & 30.0M / 25.6M & 63.0 & 74.0 & 83.4 & 90.9\\
    \midrule
    \multirow{1}{*}{None} & \multirow{1}{*}{\textbf{SimMatch (Ours)} } &
     $\sim$400 & 30.0M / 25.6M & \textbf{67.2} & \textbf{74.4} & \textbf{87.1} & \textbf{91.6} \\
	\bottomrule
	\end{tabular}
	\label{table:imagenet_result}
\end{table*}

\renewcommand\arraystretch{0.9}
\begin{table*}
    \centering
    \caption{Transfer learning performance using ResNet-50 pretrained with ImageNet. Following the evaluatiion protocal from \cite{simclr, byol}, we report Top-1 classification accuracy except Pets and Flowers for which we report mean per-class accuracy. }
    \vspace{-8pt}
    \small
    \begin{tabular}{lccccccccc}
    \toprule
        Method & Epochs & CIFAR-10 & CIFAR-100 &  Food-101 & Cars & DTD & Pets & Flowers & Mean \\ \hline
        Supervised & - & \textbf{93.6} & 78.3 & 72.3 & 66.7 & 74.9 & 91.5 & 94.7 & 81.7 \\ \hline
        SimCLR \cite{simclr} & 1000 & 90.5 & 74.4 & 72.8 & 49.3 & \textbf{75.7} & 84.6 & 92.6 & 77.1 \\
        MoCo v2 \cite{mocov2} & 800 & 92.2 & 74.6 & 72.5 & 50.5 & 74.4 & 84.6 & 90.5 & 77.0 \\
        BYOL \cite{byol} & 1000 & 91.3 & \textbf{78.4} & \textbf{75.3} & 67.8 & 75.5 & 90.4 &  \textbf{96.1} & \textbf{82.1} \\ \hline
        SimMatch (10\%)  & 400 & \textbf{93.6} & \textbf{78.4} & 71.7 & \textbf{69.7} & 75.1 & \textbf{92.8} & 93.2 & \textbf{82.1} \\
        \bottomrule
    \end{tabular}
    \vspace{-10pt}
    \label{table:transfer_result}
\end{table*}

\subsection{ImageNet-1k}
We also performed SimMatch on the large-scale ImageNet-1k dataset \cite{imagenet_cvpr09} to show the the superiority. Specifically, we test our algorithm on 1\% and 10\% settings. We follow the same label generation process as in CoMatch \cite{comatch}, where 13 and 128 labeled samples will be selected per class for 1\% and 10\% settings respectively.

\textbf{Implementation Details}. For ImageNet-1k, we adopt ResNet-50 \cite{resnet} and use a standard SGD optimizer with Nesterov momentum. We warm up the model for five epochs until it reaches the initial learning rate 0.03 and then cosine decay it to 0. We use the same set of hyper-parameters for both 1\% and 10\% settings ($\lambda_u=10, \lambda_{in}=5, t=0.1, \alpha=0.9, \tau=0.7, \mu=5, m=0.999, B=64$). We keep the past 256 steps $p^w$ for distribution alignment. We choose the student-teacher version memory buffer and test performance on the student network. For strong augmentation, we follow the same strategy in MoCo v2 \cite{mocov2}.

\textbf{Results}. We have shown the results in Table \ref{table:imagenet_result}. As we can see, with 400 epochs training, SimMatch achieves 67.2\%, and 74.4\% Top-1 accuracy on 1\% and 10\% labeled examples, which is significantly better than the previous methods. FixMatch-EMAN \cite{eman} achieves a slightly lower performance (74.0\%) on 10\% setting. However, it requires 800 epochs of self-supervised pretrain (MoCo-EMAN) where SimMatch can directly train from scratch. The most recent work PAWS \cite{paws} achieves 66.5\% and 75.5\% Top-1 accuracy on 1\% and 10\% settings with 300 epochs training.  Nevertheless, PAWS requires the multi-crops strategy \cite{swav} and $970 \times 7$ labeled examples to construct the support set. For each epoch, the actual training FLOPS of PAWS is 4 times that of SimMatch. Hence, the reported 300 epochs PAWS should have similar training FLOPS with 1200 epochs SimMatch. Due to the limited GPU resources, we cannot push this research to such a scale, but since SimMatch surpassed PAWS on 1\% setting with 1/3 training costs (400 epochs), we believe it can already demonstrate the superiority of our method.

\textbf{Transfer Learning}. We also evaluate the learned representations on multiple downstream classification tasks. We follow the linear evaluation setup described in \cite{simclr, byol}. Specifically, we trained an L2-regularized multinomial logistic regression classifier on features extracted from the frozen pretrained network (400 epochs 10\% SimMatch), then we used L-BFGS \cite{lbfgs} to optimize the softmax cross-entropy objective, and we did not apply data augmentation. We selected the best L2-regularization parameter and learning rate from validation splits and applied it to the test sets. The datasets used in this benchmark are as follows: CIFAR-10 \cite{cifar}, CIFAR-100 \cite{cifar}, Food101 \cite{food101}, Cars \cite{cars}, DTD \cite{dtd}, Pets \cite{pets}, Flowers \cite{flowers}. The results have been shown in Table \ref{table:transfer_result}. As we can see, with only 400 epochs of training, SimMatch achieves the best performance on CIFAR-10, CIFAR-100, Cars and Flowers datasets which is comparable with BYOL and significantly better than SimCLR, MoCo V2, and supervised baseline. These results further validate the representation quality of SimMatch for classification tasks.

\begin{figure*}[!t]
    \centering
    \subcaptionbox{Pseudo-Labels Accuracy}{\includegraphics[width=.28\linewidth]{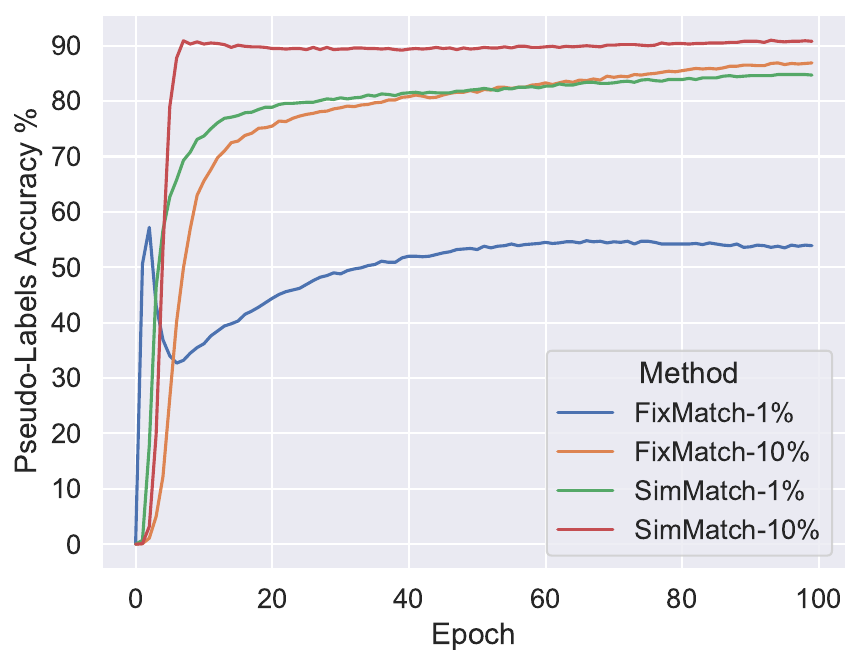}}
    \subcaptionbox{Unlabeled Samples Accuracy}{\includegraphics[width=.28\linewidth]{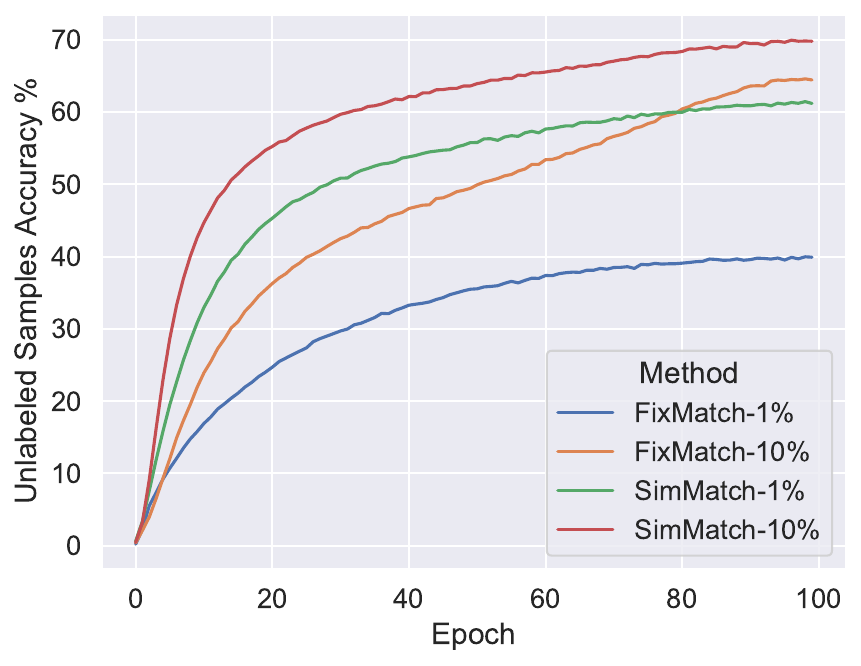}}
    \subcaptionbox{Validation Accuracy}{\includegraphics[width=.28\linewidth]{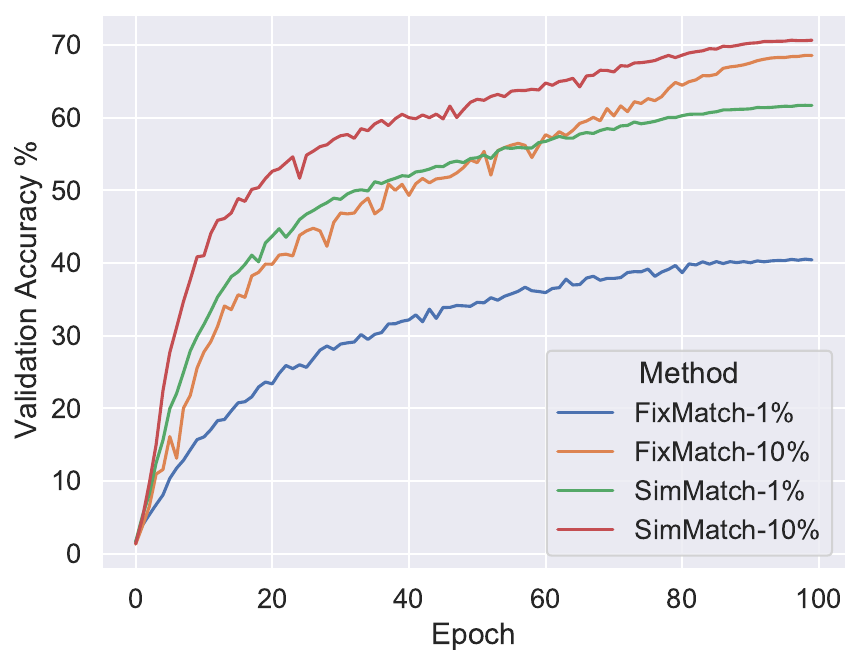}}
    \vspace{-2mm}
    \caption{Visualization of (a) pseudo-labels accuracy - the accuracy of $\widehat{p}$ that has higher confidence than threshold, (b) unlabeled samples accuracy - the accuracy of all $\widehat{p}$ regardless of the threshold, (c) validation accuracy for FixMatch and SimMatch on 1\% and 10\% setting.}
    \label{fig:comparison}
    \vspace{-4mm}
\end{figure*}

\renewcommand\arraystretch{0.8}
\begin{table}
    \centering
    \caption{GPU hours per epoch for different methods. The speed is tested on 8 NVIDIA V100 GPUs.}
    \vspace{-3mm}
    \small
    \begin{tabular}{c|c|c|c}
        Method & FixMatch & CoMatch & SimMatch (Ours)  \\ \midrule
        GPU (Hours) & 2.77  & 2.81 & \textbf{2.34}
    \end{tabular}
    \vspace{-15pt}
    \label{table:training_efficiency}
\end{table}
\textbf{Training Efficiency}. Next, we test the actual training speed for FixMatch, CoMatch, and SimMatch. The results is shown in Table \ref{table:training_efficiency}, SimMatch is nearly 17\% faster than FixMatch and CoMatch. In FixMatch, the weakly augmented $\mathcal{U}$ will be passed into the online network, which consumes more resources for the extra computational graphs. But in SimMatch, $\mathcal{U}$ only needs to be passed into the EMA network, so the computational graph does not need to be retained. Compared with CoMatch which requires two forward passes (strongly and weakly augmented $\mathcal{U}$) for the EMA network, SimMatch only requires one pass. Moreover, CoMatch adopts 4 memory banks (258M Memory) to compute the pseudo-label; SimMatch only needs 2 memory banks with 6.4M / 64M Memory for 1\% and 10\% labels, thus the pseudo-label generation will also be faster.

\subsection{Ablation Study}
\textbf{Pseudo-Label Accuracy}. Firstly, we would like to show the pseudo-label accuracy of SimMatch. In Figure \ref{fig:comparison}, we visualized the training progress of FixMatch and our method. SimMatch can always generate high-quality pseudo-labels and consistently has higher performance on both unlabeled samples and validation sets.

\begin{figure}
    \centering
    \subcaptionbox{Temperature \label{fig:ablation_t}{}}{\includegraphics[width=.4\linewidth]{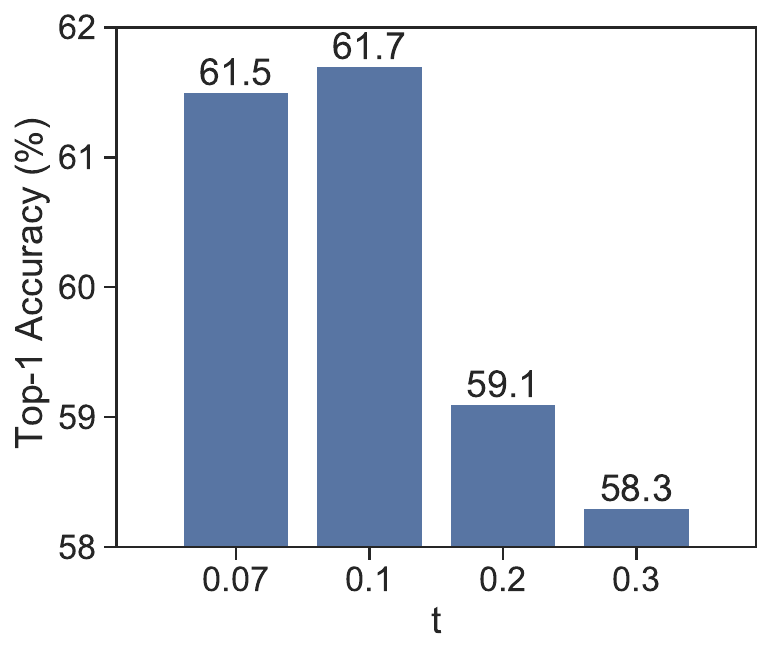}}
    \subcaptionbox{Smooth \label{fig:ablation_alpha}}{\includegraphics[width=.4\linewidth]{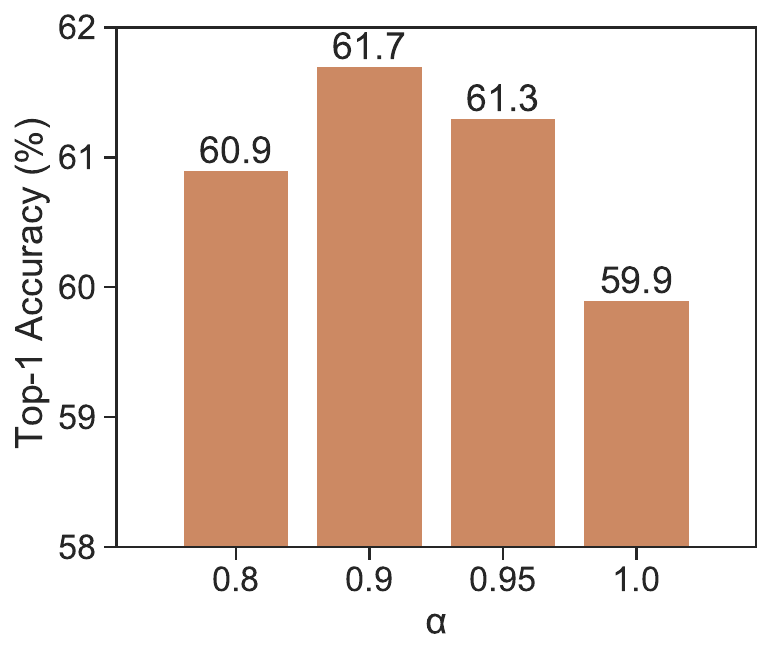}}
    \vspace{-2mm}
    \caption{Results of varying $t$ and $\alpha$. (ImageNet-1k 1\% - 100 ep) }
    \label{fig:ablation_t_alpha}
    \vspace{-5pt}
\end{figure}

\textbf{Temperature}. The temperature $t$ in Eq. \eqref{equation:student} and Eq. \eqref{equation:teacher} controls the sharpness of the instance distribution. (Noted $t=0$ is equivalent to the $argmax$ operation). We present the result of varying different $t$ value in Figure \ref{fig:ablation_t}. As can be seen, the best Top-1 accuracy comes from $t=0.1$, and slightly decreased when $t=0.07$. This is consist with the most recent works in contrastive learning where $t=0.1$ is generally the best temperature\cite{simclr, simclrv2, dino, swav}. 

\textbf{Smooth Parameter}. We also show the effective of different smooth parameter $\alpha$ Eq. \eqref{equation:calibratep} in Figure \ref{fig:ablation_alpha}. Specifically, we sweep over $[0.8, 0.9, 0.95, 1.0]$ for $\alpha$, it clear to see that $\alpha=0.9$ achieves the best result.  Noted that $\alpha=1.0$ is equivalent to directly take the original pseudo-labels $p^w$ for Eq. \eqref{equation:loss_u}, which result in $1.8\%$ performance drop. 

\textbf{Label Propagation}. Next, we would like to verify the effectiveness of the label propagation. The results has been shown in Table \ref{table:ablation_remove}. When we remove $\widehat{p}$, this is the same case with $\alpha=1.0$, so we will not discuss this setting further. If we remove $\widehat{q}$, which means the projection head will be trained in a fully unsupervised manner as in \cite{ressl}, as we can see the performance is significantly worse than standard SimMatch, demonstrating the importance of our label propagation strategy.
\renewcommand\arraystretch{0.8}
\begin{table}
    \centering
    \setlength\tabcolsep{10pt}
    \caption{Results of removing scaling and smoothing strategy. (ImageNet-1k 1\% - 100 ep)}
    \vspace{-5pt}
    \small
    \begin{tabular}{c|c|c|c}
        Method & w/o $\; \widehat{p}$ & w/o $\; \widehat{q}$ & Standard \\ \midrule
        Top-1 & 59.9 & 52.3 & \textbf{61.7} \\
    \end{tabular}
    \label{table:ablation_remove}
    \vspace{-10pt}
\end{table}

\textbf{Propagation Strategy}. Then, we tried different combinations of the scaling and smoothing strategy to generate the pseudo-labels $\widehat{p}$ and $\widehat{q}$. From Table \ref{table:ablation_strategy}, we can see that take smoothing for $\widehat{p}$ and scaling for $\widehat{q}$ achieves the best result. We might notice that applying smoothing to both $\widehat{p}$ and $\widehat{q}$ can achieve similar performance ($61.5\%$). However, the smoothing strategy will introduce a smoothing parameter. Thus, for keeping our framework simple, we prefer to choose the scaling strategy for $\widehat{q}$.
\renewcommand\arraystretch{0.85}
\begin{table}
    \centering
    \setlength\tabcolsep{15pt}
    \caption{Results of different combinations for scaling and smoothing strategy. (ImageNet-1k 1\% - 100 ep)}
    \vspace{-5pt}
    \small
    \begin{tabular}{|l|c|c|}
    \hline
    \diagbox{$\widehat{p}$}{$\widehat{q}$} & Scaling & Smoothing \\ \hline
    Scaling & 56.6 & 59.9 \\ \hline
    Smoothing & \textbf{61.7} & 61.5 \\ \hline
    \end{tabular}
    \label{table:ablation_strategy}
    \vspace{-10pt}
\end{table}

\textbf{Instance Matching Loss Design}. To verify the effectiveness of the instance similarity matching term $\mathcal{L}_{in}$, we simply replace it with InfoNCE and SwAV. We show the result in Table \ref{table:ablation_replace} When working with InfoNCE loss, we sweep the temperature over [0.07, 0.1, 0.2]. In this case, the best result we can get is 53.5\%, which is 8.2\% lower than SimMatch. This is due to the natural conflict between the classification problem and InfoNCE objective. To be specific, the classification problem aims to group similar samples together, but InfoNCE aims to distinguish every instance. When working with SwAV, we tried to set the number of prototypes to 1000, 3000, and 10000. Finally, the best result we can get is 49.7\%, which is 12\% lower than SimMatch. SwAV aims to distribute the samples equally to each prototype, preventing the model from collapsing. However, distribution alignment has a similar objective, which is adopted by SimMatch in Eq \eqref{equation:loss_u} . Moreover, the SwAV loss will be trained in a completely unsupervised manner, which will lose the power of the labels. The advantage of $\mathcal{L}_{in}$ is that the label information can easily cooperate with instance similarities.
\renewcommand\arraystretch{0.8}
\begin{table}
    \centering
    \setlength\tabcolsep{10pt}
    \caption{Results of replacing $\mathcal{L}_{in}$ with InfoNCE and SwAV. (ImageNet-1k 1\% - 100 ep)}
    \vspace{-3.5mm}
    \small
    \begin{tabular}{c|c|c|c}
        Method & InfoNCE & SwAV & SimMatch \\ \midrule
        Top-1 & 53.5 & 49.7 & \textbf{61.7} \\
    \end{tabular}
    \label{table:ablation_replace}
    \vspace{-10pt}
\end{table}

\section{Conclusion}
In this paper, we proposed a new semi-supervised learning framework SimMatch, which considers the consistency regularization on both semantic-level and instance-level. We also introduced a labeled memory buffer to fully leverage the data annotations on instance-level. Finally, our defined \textit{unfolding} and \textit{aggregation} operation allows the label to propagate between semantic-level and instance-level information.  Extensive experiment shows the effectiveness of each component in our framework. The results on ImageNet-1K demonstrate the state-of-the-art performance for semi-supervised learning.

\section*{Acknowledgment}
This work is funded by the National Key Research and Development Program of China (No. 2018AAA0100701) and the NSFC 61876095. Chang Xu was supported in part by the Australian Research Council under Projects DE180101438 and DP210101859. Shan You is supported by Beijing Postdoctoral Research Foundation.

{\small
\bibliographystyle{ieee_fullname}
\bibliography{egbib}
}

\end{document}